\documentclass[journal]{IEEEtran}
\usepackage{blindtext}
\usepackage{graphicx}
\usepackage{multirow}
\usepackage{booktabs}
\usepackage{comment}
\usepackage{xcolor}
\usepackage{amssymb}
\usepackage{mathtools}
\usepackage{hyperref}
\usepackage{tikz}
\usepackage{pgfplots}
\usepackage{color, colortbl}

\usepackage{url}

\begin{document}

\title{Global-Local Transformer \\for Brain Age Estimation}

\author{Sheng He, P. Ellen Grant, Yangming Ou
       
\thanks{
S He, PE Grant and Y Ou are with the Boston Children's Hospital and Harvard Medical School, Harvard University, 300 Longwood Ave., Boston, MA, USA (e-mail: heshengxgd@gmail.com; yangming.ou@childrens.harvard.edu)}

}

\maketitle

\begin{abstract}
Deep learning can provide rapid brain age estimation based on brain magnetic resonance imaging (MRI).
However, most studies use one neural network to extract the global information from the whole input image, ignoring the local fine-grained details.
In this paper, we propose a global-local transformer, which consists of a global-pathway to extract the global-context information from the whole input image and a local-pathway to extract the local fine-grained details from local patches.
The fine-grained information from the local patches are fused with the global-context information by the attention mechanism, inspired by the transformer, to estimate the brain age.
We evaluate the proposed method on 8 public datasets with 8,379 healthy brain MRIs with the age range of 0-97 years.
6 datasets are used for cross-validation and 2 datasets are used for evaluating the generality. 
Comparing with other state-of-the-art methods, the proposed global-local transformer reduces the mean absolute error of the estimated ages to 2.70 years and increases the correlation coefficient of the estimated age and the chronological age to 0.9853.
In addition, our proposed method provides regional information of which local patches are most informative for brain age estimation.
Our source code is available on: \url{https://github.com/shengfly/global-local-transformer}.
\end{abstract}


\IEEEpeerreviewmaketitle

\section{Introduction}
\label{sec:intro}

 Brain age can be estimated by using machine learning techniques on the brain magnetic resonance image (MRI)~\cite{armanious2021age,hu2020disentangled,feng2020estimating,cheng2021brain}.
The MRI-derived brain age is associated with brain health at the individual level~\cite{cole2017predicting,beheshti2018association,gaser2013brainage,brainwaa160}.

The difference between the predicted brain age and the chronological age is called ``brain age gap (BAG)", which is an informative biomarker of brain health~\cite{cole2017predicting}.
Many studies have shown that a positive BAG associates with a risk of cognitive decline and neurodegeneration, such as Alzheimer's disease~\cite{beheshti2018association}, mild cognitive impairment (MCI)~\cite{gaser2013brainage,brainwaa160}, mortality~\cite{cole2018brain}, psychosis~\cite{chung2018use}, major depressive disorder~\cite{han2020brain} and others~\cite{cole2017predicting,kaufmann2019common}.
The key part of brain age estimation is to train a machine learning model on a normal aging population which can estimate brain age on healthy brain MR images with low errors.
In most studies~\cite{hu2020disentangled,feng2020estimating,peng2019accurate}, the machine learning model is trained on brain MR images with the individual's chronological age which is the amount of time since the birth of an individual~\cite{armanious2021age}.
Thus, the trained machine learning model can estimate the brain age on unseen data by extracting the chronological age-specific patterns learned from the healthy brain MRIs.

Convolutional neural networks (CNNs) can provide superhuman performance on various applications~\cite{krizhevsky2012imagenet,lecun2015deep}, including brain age estimation~\cite{feng2020estimating,brainwaa160,cole2017predictingcnn,jonsson2019brain,levakov2020deep}.
It can make a prediction on  both the whole input images and local patches (segmented from the input image)~\cite{brendel2018approximating}.
One advantage of using CNNs on the whole images is that it can capture the global information and provide an image-level prediction. 
However, fine-grained details are missed due to the fact that deep neural networks are dominated by the salient information on the whole image.
In addition, the decision of the CNN from the whole input image is hard to understand~\cite{feng2020estimating,he2016deep}.
On the contrary, patch-based method can capture the local detailed information and provide a patch-wise evidence which can reveal the age-specific patterns for interpretation~\cite{brendel2018approximating,qiu2020development}.
However, the performance is limited due to the lack of the global context information.

\begin{figure}[!t]
    \centering
    \includegraphics[width=0.5\textwidth]{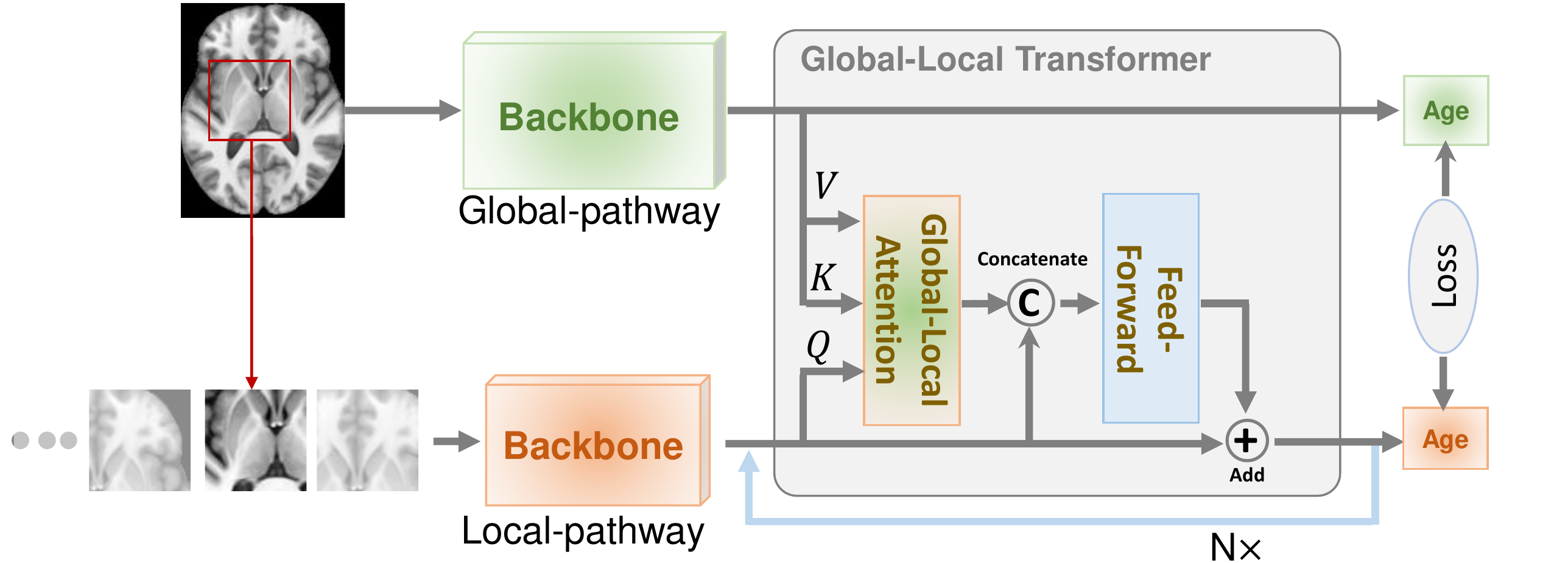}
    \caption{The proposed framework for brain age estimation, which contains two pathways: the global-pathway extracts the global-context information from the whole input image and the local-pathway extracts deep features from local patches segmented from the input image. The global-pathway and the local-pathway interact in a global-local transformer block to fuse the global-context and local fine-grained details for brain age estimation. There are $N$ identical global-local transformers to iteratively fuse the global and local features.}
    \label{fig:framework}
\end{figure}

To solve the problem, we present a two-pathway network for brain age estimation.
One pathway is designed to capture the global context information from the input brain MRI and another pathway is responsible for capturing the fine-grained information from a local patch.
We fuse the local detailed and the global context information with an attention mechanism, inspired by the self-attention in the ``transformer"~\cite{vaswani2017attention}.
Thus, our proposed fusion block is named ``global-local transformer", as shown in Fig.~\ref{fig:framework}.

\subsection{Our method exploits global context and local details.}
The global-pathway makes a decision based on the whole input image and the deep features contain global-context information of the input image.
However, it is easy to converge to the most informative regions yielding a small training loss~\cite{zhou2016learning} and other regions which also contain the subtle age information are ignored.
The local-pathway learns the age information from a local patch, forcing the network to learn the detailed age information within a small local region~\cite{brendel2018approximating}, yielding a limited performance because the size of the receptive field is bounded in the local patch.
Many studies in the literature have shown that fusing the global-context and local detailed information can improve the performance~\cite{van2021hooknet,feichtenhofer2019slowfast,chen2019collaborative,he2020fragnet,guo2019deep,yan2019learning}.
Our proposed method uses the attention mechanism to optimally fuse the global-context information extracted from the global-pathway and local detailed information extracted from the local-pathway.

\subsection{Our method does not need spatial feature alignment.}
A common way for fusing the global and local features from two different pathways is to segment features from the global pathway at the same spatial location of the local patch and concatenate them together~\cite{van2021hooknet,chen2019collaborative,he2020fragnet}.
However, there are two limitations: (1) it requires that the features from global and local pathways being spatially aligned~\cite{van2021hooknet}, which is difficult for an arbitrary input image size after several max-pooling layers in the neural network; 
(2) the cropped deep features from the global-pathway still contain information from local regions, without the global-context information.

We use the attention mechanism~\cite{vaswani2017attention,schlemper2019attention} to optimally fuse the deep features from the global-pathway and local-pathway. 
The attention can select the most important information and ignore the irrelevant information on the context features from the global-pathway.
A weighted sum of the global-context information at all positions is fused with the feature on each location of the deep features from the local-pathway, where the weight (normalized by the softmax, named attention) is computed by the similarity between the corresponding global and local deep features.
Thus, our method does not need any spatial feature alignment and can capture long-range global-context information guided by the similarities between the features from the global-pathway and local-pathway.

\subsection{Our method can be interpreted.}
There are different methods to interpret the deep learning methods for brain age estimation.
Levakov et al.~\cite{levakov2020deep} applied a gradient-based method to compute individual explanation maps which represent the contribution of each voxel to brain age prediction.
Our previous work~\cite{he2021multi} computed a correlation map between the response of the hidden neurons and the chronological ages over a population to find the most discriminative neurons in the neural network.
For the transformer, the attention flow~\cite{abnar2020quantifying} could be used to evaluate the relative relevance of the patches.
These indirect interpretation methods aim to understand where or what the neural network has learned from brain images with the limitation that the neural networks are dominated by salient information.

The direct interpretation method, on the other hand, interprets the neural network by training it directly on local patches and quantifying prediction accuracy on every local patch, to highlight the most informative patches in input images.
One representative method is the BagNet~\cite{brendel2018approximating}, which
classifies an image based on small local patches segmented from images without considering their spatial orders, making it easy to analyze the predicted evidence from each local patches.
Similar to BagNet, our method can estimate brain age based on local patches. 
Thus, the patch-level evidence for each subject can be exploited and visualized for interpretation~\cite{brendel2018approximating,qiu2020development}.
The proposed method shares the advantages of BagNet.
In addition, the performance of the neural network on local-patches is higher than BagNet since it also learns the corresponding global-context information with attention from the global-pathway.

\section{Related work}
\subsection{Brain age estimation}

Table~\ref{tab:soasummary} summaries related studies in the literature for brain age estimation using convolutional neural networks.
Most studies use the common structural networks for brain estimation on brain MRIs~\cite{huang2017age,ueda2019age,dinsdale2021learning}.
A 3D neural network with 5 convolutional layers and one fully connected layer for brain age regression was proposed in~\cite{cole2017predictingcnn}, providing the mean absolute error (MAE) of 4.16 years on the age range of 18-90 years.
The 3D version of the residual network~\cite{he2016deep} has been applied in~\cite{jonsson2019brain}, achieving the MAE of 3.631 years by combining predictions from multiple CNNs.
The 3D version of the VGG network~\cite{simonyan2014very} was employed for brain age estimation~\cite{jiang2019predicting} on subjects aged 18-90 years with the MAE of 5.55 years.
Feng et al.~\cite{feng2020estimating} used a neural network with 10 convolutional layers for brain age estimation on subjects with ages from 18-97, yielding an MAE of 4.21 years.
Levakov et al.~\cite{levakov2020deep} utilized a 3D CNN with 4 convolutional layers and 2 fully-connected layers for estimating brain age and an MAE of 3.07 years is obtained by averaging of 10 CNNs on 10,176 subjects (age range: 4-94 years).
Peng et al.~\cite{peng2019accurate} proposed a lightweight Simple Fully Convolutional Network (SFCN), achieving an MAE of 2.14 years on subjects with age range 44-80 years.
Bashyam et al.~\cite{brainwaa160} developed a DeepBrainNet for brain age prediction based on 2D slices, providing an MAE of 3.702 years on a large set of MRI scans.

Recently, Cheng et al.~\cite{cheng2021brain} proposed a 3D two-stage-age-network to estimate brain age from T1w MRI with two stages: the first stage estimated a rough brain age and the second stage was used to refine the results.
An MAE of 2.428 was achieved on 6,586 subjects with ages of 17-98 years.
Our previous work in~\cite{he2021multi} used a fusion-with-attention (FiA-Net) 3D network to fuse the intensity and RAVENS channels for brain age estimation, yielding an MAE of 3.00 years on a lifespan cohort (0-97 years).

\begin{table}[!t]
    \centering
    \caption{A summary of machine learning studies for brain age estimation using Convolutional neural networks on healthy brain MRIs. (MAE: mean absolute errors)}
    \label{tab:soasummary}
    \resizebox{0.5\textwidth}{!}{
    \begin{tabular}{l|cccc}
    \toprule
    Study & Publish Year  & \#Samples & Age ranges & MAE (years) \\
    \midrule
    Huang et al.~\cite{huang2017age} & 2017&  1,099 & 20-80 & 4.0 \\
    Cole et al.~\cite{cole2017predictingcnn}  &  2017 &  2,001 & 18-90 & 4.16 \\
    Ueda et al.~\cite{ueda2019age} & 2019 &  1,101 & 20-80 & 3.67 \\
    J{\'o}nsson et al.~\cite{jonsson2019brain} & 2019 & 1,264 & 15-80 & 3.63 \\
    Jiang et al.~\cite{jiang2019predicting} & 2020 &  1,454 & 18-90 & 5.55 \\
    Feng et al.~\cite{feng2020estimating} & 2020 &  10,158 & 18-97 & 4.06 \\
    Bashyam et al.~\cite{brainwaa160} & 2020 &  11,729 & 3-95 & 3.702 \\
    Levakov et al.~\cite{levakov2020deep} & 2020 & 10,176 & 4-94 & 3.07 \\
    Peng et al.~\cite{peng2019accurate} & 2021 &  14,503 & 42-82 & 2.14 \\
    Dinsdale et al.~\cite{dinsdale2021learning} & 2021  & 19,687 & 44-80 & 2.97 \\
    He et al.~\cite{he2021multi} & 2021 & 16,705 & 0-97 & 3.00 \\
    Chen et al.~\cite{cheng2021brain} & 2021 & 6,586 & 17-98 & 2.428 \\
    Proposed & - & 8,379 & 0-97 & 2.70 \\
    \bottomrule
    \end{tabular}}
\end{table}

Our method is different in two key ways: (1) We propose a two-pathway network, which can exploit the global-context and local detailed information for brain age estimation. (2) We apply the proposed method on 2D slices extracted from 3D brain MRI columns, which is computationally efficient and achieves the MAE of 2.70 years on a lifespan (0-97 years).

\subsection{Transformer}
Transformer~\cite{vaswani2017attention} was first used for natural language processing (NLP) and has recently become popular in visual recognition~\cite{dosovitskiy2020image,chen2020generative}.
The core idea is to apply a self-attention layer on the input sequence to capture the relationship among the sequence of local patches.
The input sequence is first converted into three different components, namely ``query", ``key" and ``value". Subsequently, the attention is obtained based on ``query" and ``key", and applied on the ``value" to output a scaled sequence.
Transformer has been used in image recognition~\cite{dosovitskiy2020image}, object detection~\cite{carion2020end}, hand pose estimation~\cite{huang2020hand}, image super-resolution~\cite{yang2020learning}, etc. 
A recent survey can be found in~\cite{han2020survey}.

Our proposed method is different in the way that  the ``query", ``key" and ``value" are from different features: we compute the ``key" and ``value" from the global-pathway and the ``query" from the local-pathway.
Through the ``key" and ``query", the attention between the the global and local information can be obtained and applied to the ``value" to compute the global-context information for the local patches. 
Thus, our method can optimally fuse the global-context and local detailed information with attention, which is named ``global-local attention".
The corresponding transformer with ``global-local attention" is named ``global-local transformer".

\section{Method}
\subsection{Backbone for deep feature extraction}
\label{sec:vgg}

\begin{figure}
    \centering
    \includegraphics[width=0.5\textwidth]{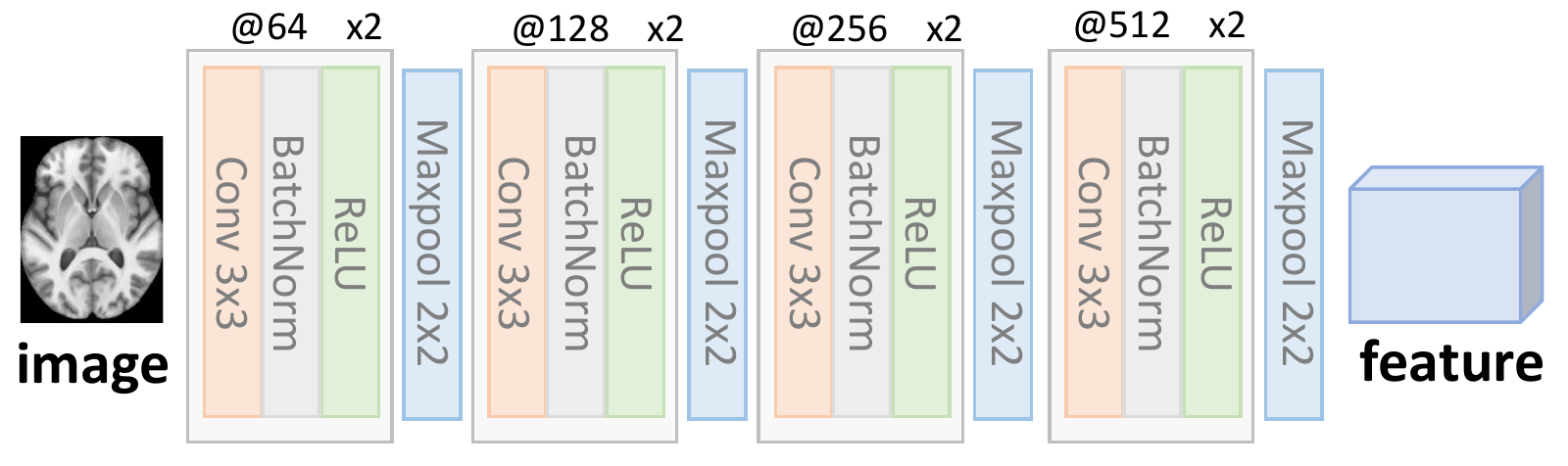}
    \caption{The backbone of the convolutional neural network. It takes brain image as input and converts it to a deep feature. The network contains 8 blocks and each block has one convolutional layer, one batch normalization layer and one ReLU layer. The spatial resolution is reduced by the max-pooling layer. (@n denotes n channels and x2 means there are two stacked identical blocks).}
    \label{fig:backbone}
\end{figure}

We use a convolutional neural network (CNN) as the backbone to the extract deep features from the input image.
The backbone is based on VGGNet~\cite{simonyan2014very} but with a small number of layers based on the fact that ``shallow neural networks provide better results than deep ones in brain age estimation~\cite{peng2019accurate}".
As show in Fig.~\ref{fig:backbone}, the backbone contains eight blocks.
Each block consists of a convolutional layer with kernel size of 3$\times$3 and padding of 1, a batch normalization layer~\cite{ioffe2015batch} and a ReLU activation layer~\cite{lecun2015deep}.
A max-pooling layer with the kernel size of 2$\times$2 and stride of 2 is applied after every two blocks to gradually reduce the spatial dimensions.
The channel numbers used in each block are [64,128,256,512], similar to VGGNet~\cite{simonyan2014very}.
The backbone converts an input image into a deep feature, representing the abstract and high-level features of the input image.

\subsection{Global-local attention mechanism}
\begin{figure}[!t]
    \centering
    \includegraphics[width=0.5\textwidth]{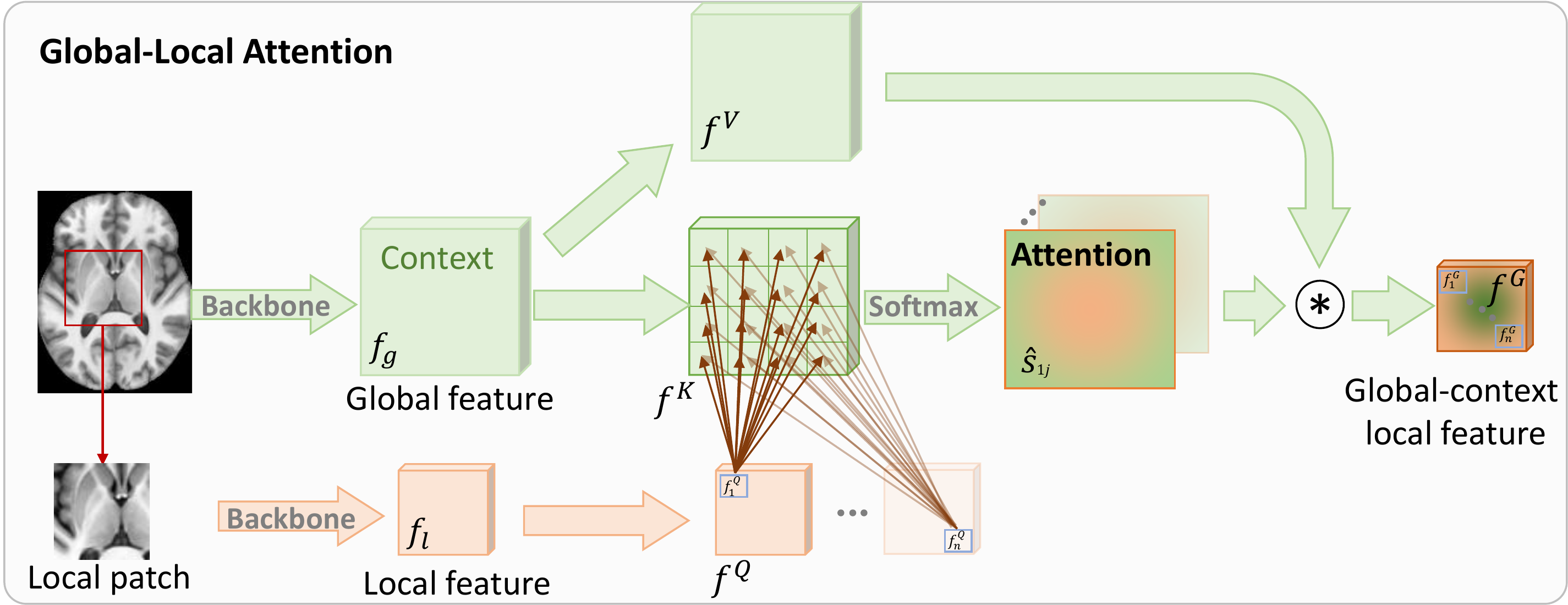}
    \caption{The framework of global-location attention mechanism.  The attention is computed at every pixel location. Each grid in $f^K$ and $f^Q$ represents one pixel for improving the visualization. }
    \label{fig:glattention}
\end{figure}

Our aim is to develop a neural network to learn the fine-grained information from a local patch by fusing the global-context visual information learned from the whole input image.
We use two identical backbones with different parameters to extract deep feature $f^{d\times h\times w}_g$ from the whole input image (with height $h$ and width $w$) and deep feature $f^{d\times h'\times w'}_l$ from a local patch (with height $h'$ and width $w'$) with the number of channels $d$, the height $h'<h$ and the width $w'<w$.
The deep feature $f_g$ contains the global-context information of the whole input image and the deep feature $f_l$ contains the fine-grained local information from a local patch.
To fuse the global-context and local fine-grained information, we propose a global-local attention, inspired by the self-attention mechanism~\cite{vaswani2017attention}.
The framework is shown in Fig.~\ref{fig:glattention}.
For the local feature $f_l$, we  use a $1\times 1$ convolutional layer to project it into a new space $f^Q$, named ``query".
For the global feature $f_g$, we use two $1\times 1$ convolutional layers to project it into two different spaces $f^K$ and $f^V$, named ``key" and ``value".

As shown in Fig.~\ref{fig:glattention}, for the deep feature at each location (pixel-level) of the query $f^Q_i$ (where $i$ is the position index on the local feature), we compute its similarity to all positions on the deep feature $f^K$, using the dot-product function:
$s_{ij} =  f^Q_i\cdot f^K_j$ (where $j$ is the position index on the global features $f^K$). 
We divide the $s_{ij}$ by $\sqrt{d}$ and normalize it by applying a softmax function:
$\hat{s}_{ij}=\text{softmax}_j(s_{ij}/\sqrt{d})$.
The normalized $\hat{s}_{ij}$ is called attention since it has different weights on different locations $j$, determined by both the query $f^Q_i$ and the key $f^K_j$.
Finally, the corresponding global-context information for each location on the local feature $f^Q_i$ is obtained as the sum of the weighted value computed from the global-pathway: $f^G_i=\sum_j \hat{s}_{ij}\cdot f^V_j$.

Similar to the self-attention~\cite{vaswani2017attention}, we can also use the matrix operations to efficiently compute the global-local attention.
We reshape the global features $f^K$ and $f^V$ and local feature $f^Q$ to $F^K$ (with the size of $N_1\times d, N_1=h\times w)$, $F^V$ ($N_1\times d)$ and $F^Q$ ($N_2\times d, N_2=h'\times w')$, respectively. 
The global-context local feature can be defined by:
\begin{equation}
\label{eq:attention}
    F^G=\text{softmax}(\frac{F^Q(F^K)^T}{\sqrt{d}})F^V
\end{equation}

The global-local attention is different from the self-attention~\cite{vaswani2017attention} in two different aspects.
First, in self-attention, the ``query", ``key" and ``value" are from the same feature vectors while in the global-local attention, the ``query" ($F^Q$) is from the local pathway and the pair of ``key" ($F^K$) and ``value" ($F^V$) are from the global pathway.
Second, in self-attention, the ``query", ``key" and ``value" usually have the same size while in the global-local attention, the size of $F^Q$ is smaller than the size of $F^K$ and $F^V$ because $F^Q$ is computed from a small local path.
The global-local attention is asymmetry: while $F^K$ and $F^V$ are derived from the global-pathway, $F^Q$ is derived from the the local-pathway, with the number of features ($N_1\ll N_2$) and the complexity of the global-local attention (as defined in Eq.~\ref{eq:attention}) is $\mathcal{O}(N_1N_2)$.

As show in Fig.~\ref{fig:glattention}, the output size of the global-context local feature $F^G$ is the same as the size of the local feature $F^Q$.
The feature values on the output are computed as a weighted sum of the values $F^V$ from the global pathway.
Thus, the output contains the global contextual information determined by the global and local features without any spatial alignment. 

Similar to the self-attention, we also use multi-head attention, where the global and local features are split into $h=8$ parallel parts on the channel dimension. 
The global-local attention is applied on each part and the output values are concatenated and projected into one feature with the same size of the input feature. 
Multi-head attention becomes a standard component in Transformer~\cite{kentonbertbertpre} and more details can be found in the self-attention literature~\cite{vaswani2017attention}.

\subsection{Global-local Transformer}
In this section, we present the global-local transformer, as shown in Fig.~\ref{fig:framework}.
We concatenate the output of the global-local attention block with the local feature because it contains the global contextual information from the global pathway which is different from the local feature.
The global-context information is a weighted sum (with attention) of the global features according to the similarity determined by both global and local features. 
These two different features are further fused in a feed-forward block.
Slightly different from the standard transformer~\cite{vaswani2017attention},  the feed-forward block contains two linear transformations (two convolutional layers with 512 channels and a kernel size of 1) with batch normalization and ReLU activation functions to fuse the global contextual and local fine-grained information. 
The output is added with the local feature inspired by the residual learning~\cite{he2016deep}.
Thus, it is the local features that contain the global-context information.
The global-context information and the updated local feature can be also fed into another global-local transformer.
The same structure is repeated $N$ times to iteratively integrate the global-context and local detailed information (the selection of $N$ will be discussed in Section~\ref{sec:evaluation}).
The final layer on each branch for brain age estimation is a fully-connected layer to map the feature vector (obtained after an average pooling layer) with 512 dimensions to the brain age.

\section{Experiments}
In this section, we present the experimental results of the proposed method on a large healthy cohort. 
We also compare it with baseline models and state-of-the-art architecture of neural networks. 

\subsection{Dataset}

\begin{table}[!t]
    \centering
    \caption{The information of datasets in the healthy cohort used for brain age estimation.}
    \label{tab:dataset}
    \begin{tabular}{l|ccc}
    \toprule
    Dataset & $N_{\text{samples}}$ & Age range & Gender(female/male) \\
    \midrule
    $\Diamond$ BGSP~\cite{holmes2015brain} & 1,570 & 19.0-35.0 & 905/665\\
    $\Diamond$ OASIS-3~\cite{lamontagne2018oasis}   &  1,222 & 42.0-97.0 & 750/472\\
    $\Diamond$ NIH-PD~\cite{evans2006nih}     &  1,211 & 0-22.2 & 626/585\\
    $\Diamond$ ABIDE-I~\cite{di2014autism} & 567 & 6.4-56.2 & 98/469 \\
    $\Diamond$ IXI* & 556 & 19.9-86.3 & 309/247 \\
    $\Diamond$ DLBS~\cite{park2012neural} & 315 & 20.5-89.1 & 198/117 \\
    \midrule
    $\oplus$ CMI~\cite{alexander2017open} & 1,765 & 5.0-21.9 & 1,117/648 \\
    $\oplus$ CoRR~\cite{zuo2014open} & 1,173 & 6.0-88.0 & 591/582 \\
    \midrule
    Overall & 8,379 & 0-97.0 & 4,594/3,785\\
    \bottomrule
    \multicolumn{4}{l}{*\url{https://brain-development.org/ixi-dataset/}} \\
    \multicolumn{4}{l}{$\Diamond$ Datasets are used for cross-validation. } \\
    \multicolumn{4}{l}{$\oplus$ Datasets are used for evaluating the generality. } \\
    \end{tabular}
\end{table}

In this paper, we evaluate the proposed method on a healthy cohort:
we collect the healthy brain T1-weighted MRI scans from 8 public data sets (Table~\ref{tab:dataset}), with a total of 8,379 samples with an age range of 0-97 years.
Among of them, 6 data sets are used for cross-validation and the CMI and CoRR data sets are used for evaluating the generality of the deep learning models.

The pre-processing steps include N4 bias correction~\cite{tustison2010n4itk}, field of view normalization~\cite{ou2018field}, and Multi-Atlas Skull Stripping (MASS)~\cite{doshi2013multi}. 
The skull-stripped T1w MRI is affine registered to the SRI atlas \cite{rohlfing2010sri24} by FSL's flirt tool~\cite{jenkinson2001global}, which has a voxel size of $1\times1\times1 mm$, and was constructed from T1w of 24 healthy brains. 
The dimension of the 3D brain volume is cropped into the size of 130$\times$170$\times$120 by removing the black boundaries.
All MRI scans have been checked manually to remove the failure MRIs with severe artifacts or poor registration.

We extract 2D slices around the center of the 3D brain volumes in the axial, coronal, and sagittal planes, a strategy similar to that in~\cite{brainwaa160,lin2018convolutional}. 
The number of 2D slices to be extracted is studied in Fig.~\ref{fig:abluation} as a key variable of our algorithm. As shown in~\cite{feng2020estimating,brainwaa160,lin2018convolutional}, 2D slices around the center of the brain in different plains can be used for brain age prediction.
In addition, training the 2D neural network requires less parameters compared to the 3D neural network.
In addition, the global-local attention is computed among every positions between the global and local features, which requires a large computational resources (computing time and memory) for the 3D neural network.
 As shown in Table~\ref{tab:dataset}, images from the BGSP, OASIS-3, NIH-PD, ABIDE-I, IXI and DLBS are randomly split into 5 parts and 5-fold cross-validation is implemented for evaluation and images scanned with different scanners from the CMI and CoRR are used for evaluating the generality of models.

\subsection{Network training}
We use the mean absolute error as the loss function, which is defined as:
\begin{equation}
\label{eq:mae}
    \mathcal{L} = \frac{1}{n}\sum_{i=1}^n | p_i - \hat{p_i} |
\end{equation}
where $p_i$ is the known chronological age of the subject and $\hat{p_i}$ is the estimated brain age from the neural network.
The final training loss is the sum of the losses form the global pathway and the local pathway.
The network is trained by the Adam optimizer built in PyTorch platform, with an initial learning rate of 0.0001, reducing to half at every 25 epochs in the total 80 training epochs.
The batch size is set to 18 due to the limitation of the GPU memory.
 The training of the neural network takes around 12 hours on a single NVIDIA RTX 6000 GPU with 12G memory.

\subsection{Performance evaluation of age estimation}
To evaluate the performance of the age estimation, we use three metrics: mean absolute error (MAE), correlation coefficient ($r$) and cumulative score (CS)~\cite{geng2007automatic}.
The MAE is defined in Eq.~\ref{eq:mae}, which is a widely used metric for brain age estimation~\cite{feng2020estimating,peng2019accurate,cole2015prediction}.
The correlation coefficient ($r$)~\cite{hu2020disentangled} is computed as the Pearson correlation between the predicted ages and the chronological ages.
The CS is the accuracy of age estimation given a threshold $\alpha$, which is given by:
\begin{equation}
    \text{CS}(\alpha)=N_{e\leq\alpha}/N\times 100\%
\end{equation}
where $N_{e\leq\alpha}$ is the number of samples on which the absolute error of age estimation $e$ is no higher than the threshold $\alpha$. 
A higher CS score means better performance.

\subsection{Comparison with different baseline models}
We compare the proposed global-local transformers with the following six different baseline models:
(1) ResNet18~\cite{he2016deep}: We train a standard ResNet with 18 layers to estimate the brain age directly on the whole input image.
(2) BagNet-ResNet18~\cite{brendel2018approximating}: The ResNet18 is applied on each local patch segmented from the input image, inspired by the BagNet~\cite{brendel2018approximating}.
(3) VGG~\cite{simonyan2014very}: We use the VGG backbone (described on Section~\ref{sec:vgg}) as the network for brain age estimation on the whole input image.
(4) BagNet-VGG~\cite{brendel2018approximating}: Applying the VGG backbone network on each local patch. 
This is similar to the BagNet-ResNet18 model by replacing ResNet18 with VGG.
(5) Global-Transformer~\cite{dosovitskiy2020image}: we use the VGG backbone to extract the feature vectors from the sequence of local patches cropped from the input image and feed the corresponding feature sequence into a standard transformer for brain age estimation.
The ``query", ``key", and ``value" are from the sequences of local patches segmented on the whole input image.
(6) Local-Transformer~\cite{vaswani2017attention}: The standard transformer is applied on the feature vectors extracted on each local patch.
The ``query", ``key", and ``value" are from the deep features extracted from the single local patch.
All models are trained with the same training configurations for fair comparison.

\subsection{Comparison with state-of-the-art neural networks}
We also compare the proposed method with other popular neural networks for visual recognition with the whole image as the input.
The compared network structures include: 
(1) ResNet50 and ResNet101~\cite{he2016deep}: the most popular residual networks with 50 and 101 layers.
(2) WRN-50 and WRN-101~\cite{zagoruyko2016wide}: Wide residual networks (WRNs) with different layers which decrease depth and increase width of residual networks.
(3) DenseNet121 and DenseNet201~\cite{huang2017densely}: Densely connected convolutional networks with different layers.
(4) SqueezeNet~\cite{iandola2016squeezenet} and ShuffleNet v2~\cite{ma2018shufflenet}: two efficient networks using  small kernel sizes or depth separable convolutional layers for visual recognition.

\subsection{Comparison with state-of-the-art brain age estimation methods.}
As we mentioned above, most studies of brain age estimation use the common structure of neural networks ( compared in the previous sections).
There are three recently published neural networks specifically designed for brain age estimation.
Thus, we also compare the proposed method with them: SFCN~\cite{peng2019accurate}, DeepBrainNet~\cite{brainwaa160} and FiA-Net~\cite{he2021multi}.
The SFCN is originally designed on 3D images, named SFCN 3D.
In addition, we replace the 3D convolutional kernels with 2D ones, named SFCN 2D, to compare the performance with 2D and 3D images.
It contains seven convolutional, batch normalization, activation and max pooling layers.
The DeepBrainNet works on 2D slices based on Inception-Res-V2~\cite{szegedy2017inception} model.
The same training configuration is applied on all these models for fair comparison.
The results of FiA-Net is directly obtained from the original paper~\cite{he2021multi} since the experimental configuration is similar to the proposed method.

\subsection{Performance of single patch size for comparison}
We evaluate the brain age performance of the proposed method with a fixed local patch size.
Although our method can segment the local patches at any position without feature alignment, we use a sliding window strategy to crop patches with a step which is set to the half of the patch size for computational efficiency.
The final estimated age is the average of the estimated ages from all possible local patches.

\subsection{Interpretation with multiple patch sizes}

\begin{figure}[!t]
    \centering
    \includegraphics[width=0.5\textwidth]{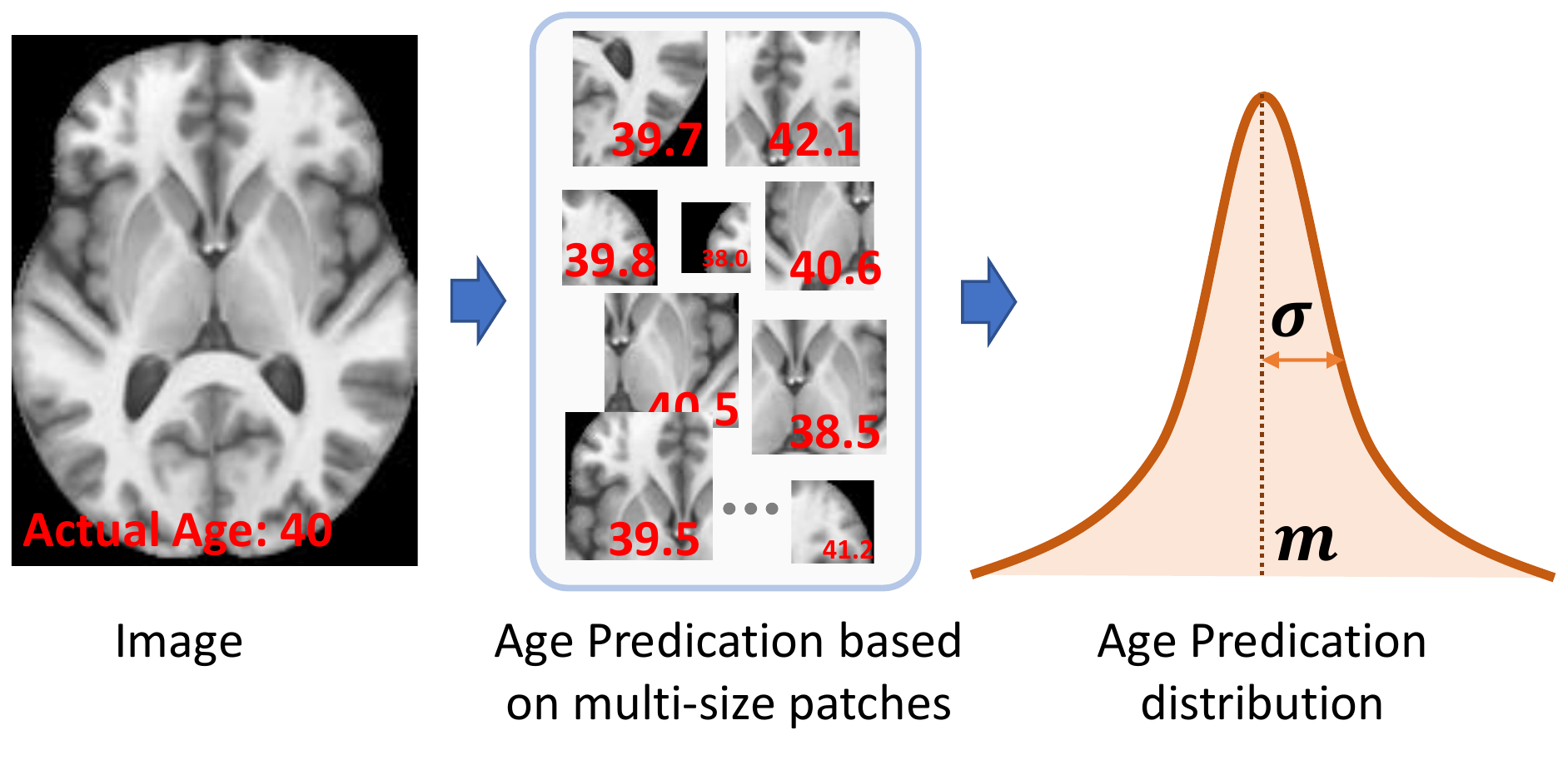}
    \caption{Illustration of brain age estimation on multiple patch sizes. Patches with different sizes are cropped from the input image and fed into the global-local transformer for age estimation. We use the mean ($m$) and standard deviation ($\sigma$) to describe the distribution of estimated ages from all possible patches.}
    \label{fig:multipatchsize}
\end{figure}

We crop the local patches with different sizes and feed them into the same local-pathway of the proposed model for brain age estimation.
In other words, all patches with different sizes share the same local-pathway in the network.
Although arbitrary patch sizes can be applied, we set the minimum patch size to 32 and maximum size to 102 with a step of 8 for computational efficiency.
During training, 30 patches are randomly sampled with different patch sizes on different locations from the whole image to train the proposed neural network.
During testing, for each subject, we randomly sample 3,000 patches and an estimated brain age is obtained on each patch.
Thus, a distribution of the estimated brain age can be obtained, described by the mean $m$ and standard 
deviation $\sigma$ (as shown in Fig.~\ref{fig:multipatchsize}).
The standard deviation $\sigma$ can be considered as the uncertainty~\cite{edupuganti2020uncertainty} of brain age estimation, which measures how differences of the estimated age on different brain regions. 
Since the brain age can be estimated on local patches, the patches with the lowest MAE can be found and visualized for interpretation.

\section{Results}
In this section, 
We first evaluate the performance of the proposed method with different parameters, such as the patch size of the local-pathway, the number of slices and global-local transformers.
In the second and third part, we compare the proposed method with different baseline models and state-of-the-art architectures.
In the last part, we visualize the most informative patches for brain age estimation.

\subsection{Parameter evaluation of the system}
\label{sec:evaluation}

\begin{figure}[!t]
    \centering
    \includegraphics[width=0.45\textwidth]{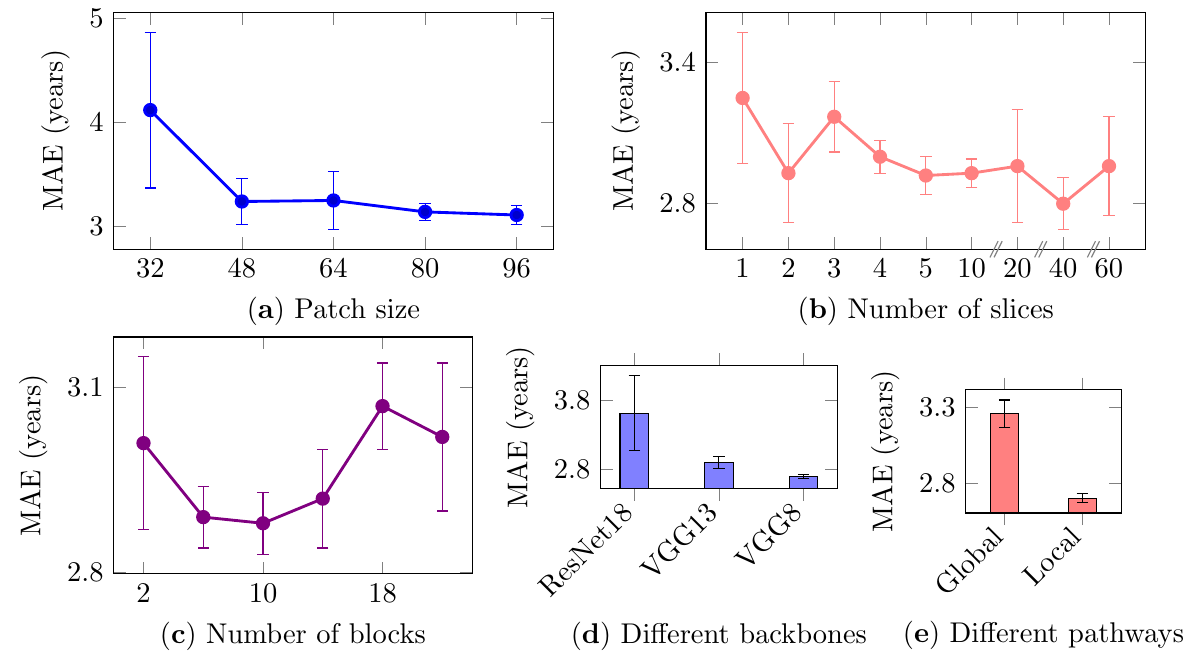}
    \caption{Performance of brain age estimation with different parameters: (a) different patch sizes; (b) different number of slices; (c) different number of blocks; (d) different backbones; (e) the global and local pathways.}
    \label{fig:abluation}
\end{figure}

First, we evaluate the performance of the proposed method with different sizes of the local patches and the experimental results are shown in Fig.~\ref{fig:abluation}(a).
From the figure we can see that there is no significant differences among results when the patch size is greater than 48. 
Thus, we set the patch size of the local-pathway to 64 in this section.
Second, we give the brain age estimation results (Fig.~\ref{fig:abluation}(b)) from 2D images with a different number of slices segmented from the 3D MRI scans, based on the fact that estimated age is closed to the actual age on the slices from the center~\cite{feng2020estimating}. 
Fig.~\ref{fig:abluation}(b) shows that there is no significant difference of performance with the number of slices from 5 to 20.
Third, we also present the performance of the system (in Fig.~\ref{fig:abluation}(c)) with a different number of global-local transformer blocks (as shown in Fig.~\ref{fig:framework}), which shows that the MAE is lower when the number of global-local transformer blocks is around N=6-10. 
In the following section, we set the number of slices to 5 and number of blocks to $N=6$ for a tradeoff between the performance and computational times and memories.
Fig.~\ref{fig:abluation}(d) shows the performance of the proposed method with different backbones: ResNet18, VGG13 and VGG8. VGG13 has a similar structure to the VGG8 (as shown in Fig.~\ref{fig:backbone}) but with 13 convolutional layers which are the same as VGG16~\cite{simonyan2014very}.
Using the neural network with 8 convolutional layers provides the best results.
A similar finding was reported in~\cite{peng2019accurate}: ``the deeper neural networks do not outperform the shallow ones in brain age prediction".
Fig.~\ref{fig:abluation}(e) shows the performance of the global and local pathways of the proposed global-local transformer.
The prediction of the local pathway is the average predicted age from all the local patches.
The local pathway captures the detailed information from the local patches with the global-context information from the global pathway, yielding a better performance than the global pathway which extracts the brain age from the whole input images.
Thus, we only report the performance of the local pathway in the following sections and the global-pathway is only used to enhance the global-context information for improving the performance of the local pathway.

\subsection{Comparison with different baseline models}

\begin{table*}[!t]
    \centering
    \caption{Performance comparison with different baseline models on three different planes and their fusion.}
    \label{tab:performace}
    \resizebox{\textwidth}{!}{
    \begin{tabular}{l|cccc|cccc}
    \toprule
    \multirow{2}{*}{Method} &  \multicolumn{4}{c|}{MAE (years)} & \multicolumn{4}{c}{Pearson correlation ($r$)}\\
    \cline{2-9}
    & Axial & Coronal & Sagittal & Fusion & Axial & Coronal & Sagittal & Fusion  \\
    \midrule
    ResNet18~\cite{he2016deep} & 3.45$\pm$0.17 & 3.38$\pm$0.07&3.87$\pm$0.10& 3.25$\pm$0.09 & 0.9759$\pm$0.0032 & 0.9764$\pm$0.0023 & 0.9700$\pm$0.0025 & 0.9789$\pm$0.0023\\
    BagNet-ResNet18~\cite{brendel2018approximating} & 5.13$\pm$0.21 & 6.29$\pm$0.81 &  5.57$\pm$0.17 & 5.46$\pm$0.34 & 0.9698$\pm$0.0026 & 0.9564$\pm$0.0065 & 0.9607$\pm$0.0029 & 0.9696$\pm$0.0028\\
    VGG~\cite{simonyan2014very}  &  3.48$\pm$0.16 & 3.29$\pm$0.09&3.67$\pm$0.13&3.11$\pm$0.12 & 0.9767$\pm$0.0024 &0.9782$\pm$0.0026 &0.9758$\pm$0.0016 &0.9801$\pm$0.0022\\
    BagNet-VGG~\cite{brendel2018approximating} & 4.08$\pm$0.09& 5.87$\pm$0.25&5.11$\pm$0.32&4.75$\pm$0.19 & 0.9758$\pm$0.0016 &0.9723$\pm$0.0016 &0.9749$\pm$0.0006 &0.9763$\pm$0.0016\\
    Global-Transformer~\cite{dosovitskiy2020image} & 3.97$\pm$0.26&5.39$\pm$0.71&5.34$\pm$0.22&4.11$\pm$0.29 & 0.9749$\pm$0.0006&0.9725$\pm$0.0036 &0.9685$\pm$0.0054& 0.9811$\pm$0.0018 \\
    Local-Transformer~\cite{vaswani2017attention}   & 3.50$\pm$0.10&3.32$\pm$0.10&3.78$\pm$0.13&3.28$\pm$0.03 & 0.9792$\pm$0.0013 &0.9774$\pm$0.0019 &0.9723$\pm$0.0020 & 0.9803$\pm$0.0015\\
    Global-Local Transformer & \textbf{2.87$\pm$0.11}&\textbf{2.97$\pm$0.07}&\textbf{3.14$\pm$0.02}&\textbf{2.70$\pm$0.03}&\textbf{0.9827$\pm$0.0022} &\textbf{0.9826$\pm$0.0022}&\textbf{0.9807$\pm$0.0019}& \textbf{0.9853$\pm$0.0020}\\
    \bottomrule
    \end{tabular}}
\end{table*}

\begin{figure}[!t]
    \centering
    \includegraphics[width=0.5\textwidth]{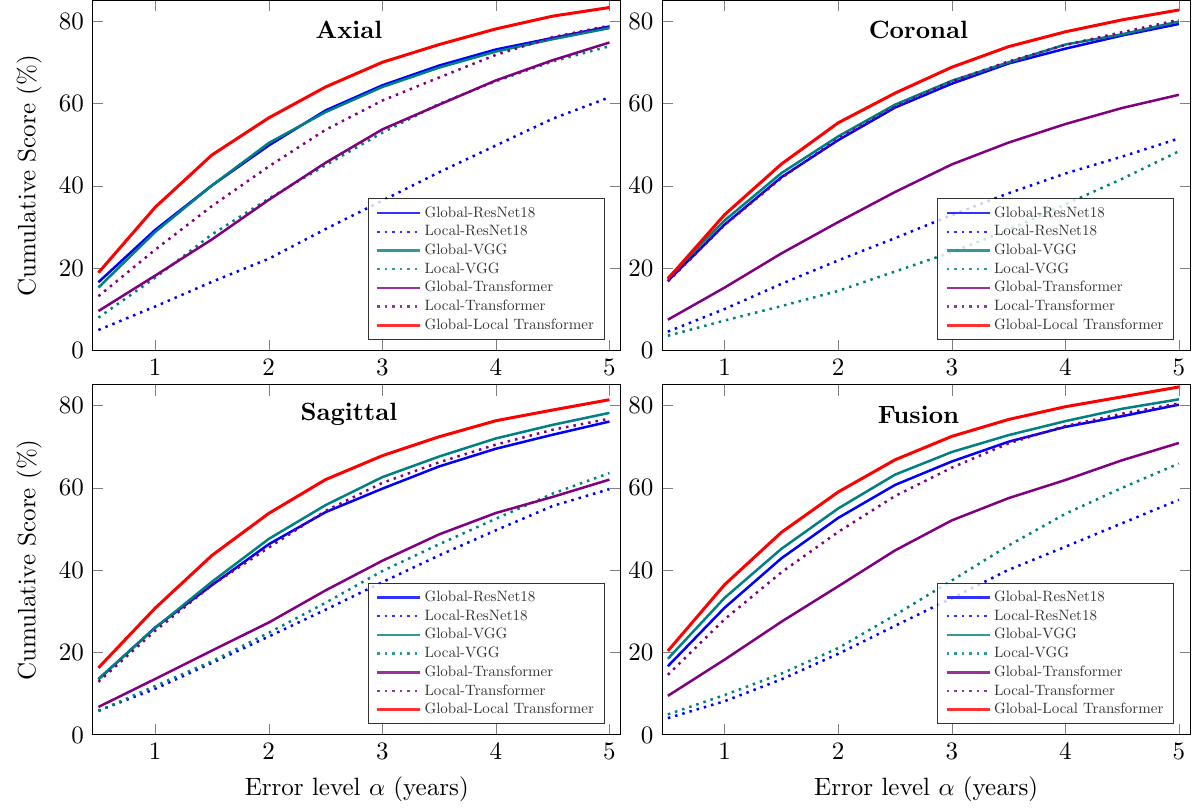}
    \caption{The Cumulative Score (CS) of error level $\alpha$ from 0 to 5 years of different models on three planes and their fusion.}
    \label{fig:CScurve}
\end{figure}

\begin{figure}[!t]
    \centering
    \includegraphics[width=0.5\textwidth]{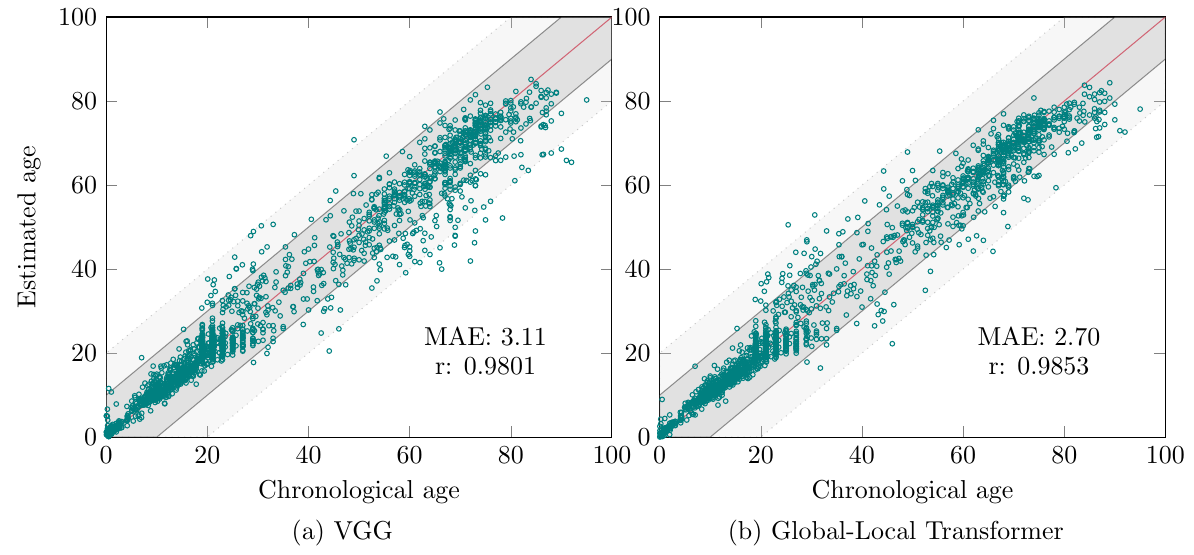}
    \caption{The scatter plots of the estimated brain ages and chronological ages based on the two best models: (a) VGG~\cite{simonyan2014very} and (b) the proposed Global-Local Transformer.}
    \label{fig:ScatterPlots}
\end{figure}

Table~\ref{tab:performace} shows the performance of different models on 2D slices extracted from the three planes: Axial, Coronal and Sagittal. 
We also fuse the results from the prediction of the 2D slices extracted from three planes by averaging the estimated ages: $y=\sum_iy_i/3$ where $y_i$ is the predicted age from the plane $i\in\{\text{Axial, Coronal, Sagittal}\}$.
Fig.~\ref{fig:CScurve} shows the CS curves of different models with different error levels $\alpha$.
Two best models' scatter plots of the estimated brain ages against the chronological ages are shown in Fig.~\ref{fig:ScatterPlots}.

Several observations can be obtained: (1) For BagNet based methods (BagNet-ResNet18 and BagNet-VGG), their performances are lower than the networks (ResNet18 and VGG) with the whole image as the input.
It shows that performance of estimating brain age based on local patches only is limited.
Using the self-attention mechanism, Local-Transformer with the VGG backbone can improve the performance, but the result is still lower than ResNet18.
In general, neural networks with the local patches as input provide lower performance than ones with the whole image as input.
However, our proposed Global-Local Transformer gives the best performance, which demonstrates the advantage of fusing the global-context and local detailed information.
(2) The age information on different planes is slightly different.
The most informative plane is the Axial, which provides better results than Cornoal and Sagittal.
The 2D slices from the Axial plane are also used for brain age estimation in~\cite{brainwaa160,armanious2021age}.
For ResNet18, VGG, Local-Transformer and the proposed Global-Local Transformer, fusing the three planes can improve the performance. 
In the following sections, we only report the performance of the fusion from the three planes since it can provide better results than the single plane.
(3) The lightweight VGG network with 8 layers provides better results than ResNet with 18 layers on both whole input image and local patches.
This is the same as the finding in~\cite{peng2019accurate} that the lightweight network can achieve better performance than ResNet~\cite{he2016deep} for brain age estimation.
(4) Our proposed Global-Local Transformer gives the lowest MAEs, highest correlation $r$ and CS with various threshold $\alpha$ than all other models on three planes and the corresponding fusion one.

To further show the detailed estimation performance of different models, the evaluation based on MAE are broken down into different age ranges.
Table~\ref{tab:agegrouperr} shows the performance on age groups which are roughly divided into four groups. 
For all models, the MAEs of estimated ages on subjects with age of 30-60 years are higher than subjects on other age groups, demonstrating the age estimation on this age group is more challenge than other age groups.
In general, the results from the table show that our proposed method always provides the best performance than all other six baseline models on four different age groups.

\begin{table}[!t]
    \centering
    \caption{Performance in different age ranges in terms of mean absolute error (MAE).}
    \label{tab:agegrouperr}
    \resizebox{0.5\textwidth}{!}{
    \begin{tabular}{l|cccc}
    \toprule
    Age group & $<$ 10 years & 10-30 years & 30-60 years & $>$ 60 years \\
    $N_{\text{samples}}$ & 690 & 2,724 & 732 & 1,296 \\
    \midrule
    ResNet18~\cite{he2016deep}   &1.39$\pm$0.21&2.23$\pm$0.11&5.88$\pm$0.25&4.90$\pm$0.15\\
    BagNet-ResNet18~\cite{brendel2018approximating}  &4.16$\pm$0.47&5.21$\pm$0.85&5.77$\pm$0.45&6.47$\pm$0.35\\
    VGG~\cite{simonyan2014very}  &1.20$\pm$0.15&2.00$\pm$0.10&5.45$\pm$0.29&5.12$\pm$0.31\\
    BagNet-VGG~\cite{brendel2018approximating} &4.26$\pm$0.44&4.30$\pm$0.38&5.23$\pm$0.19&5.64$\pm$0.25\\
    Global-Transformer~\cite{dosovitskiy2020image}  &1.82$\pm$0.29&3.29$\pm$0.36&7.16$\pm$0.63&5.31$\pm$0.70\\
    Local-Transformer~\cite{vaswani2017attention} &1.75$\pm$0.18&2.54$\pm$0.09&5.53$\pm$0.42&4.35$\pm$0.24 \\
    Global-Local Transformer &\textbf{0.97$\pm$0.12}&\textbf{1.89$\pm$0.15}&\textbf{5.12$\pm$0.34}&\textbf{3.93$\pm$0.29}\\
    \bottomrule
    \end{tabular}}
\end{table}

Table~\ref{tab:datasetperformace} shows the performance on different datasets, including the 6 datasets for cross-validation and 2 datasets for generality.
Our proposed method provides the lowest MAEs on the 6 data sets involved in the cross-validation as well as on the CMI   and CoRR data sets.
Our proposed method is generalizable to different datasets from different sites and scanners.

\begin{table*}[!t]
    \centering
    \caption{Performance on each dataset in terms of mean absolute error (MAE).}
    \label{tab:datasetperformace}
    \begin{tabular}{l|c|c|c|c|c|c|c|c}
    \toprule
    \multirow{2}{*}{Dataset} & \multicolumn{6}{c|}{Cross-validation} & \multicolumn{2}{c}{Generality} \\
    \cline{2-9}
     & BGSP & OASIS-3 & NIH-PD & ABIDE-I & IXI & DLBS & CMI & CoRR\\
    $N_{\text{samples}}$ & 1,570 & 1,222 & 1,211 & 567 & 556 & 315 & 1,765 & 1,173 \\
    Age range & 19.0-35.0 & 42.0-97.0 & 0-22.2 & 6.4-56.2 & 19.9-86.3 & 20.5-89.1 & 5.0-21.1 & 6.0-88.0 \\
    \midrule
    ResNet18~\cite{he2016deep}         & 2.00$\pm$0.03 & 3.88$\pm$0.22 & 1.16$\pm$0.11 & 3.61$\pm$0.22 & 7.98$\pm$0.87 & 6.11$\pm$0.30 & 5.34$\pm$0.25 & 6.88$\pm$0.38\\
    BagNet-ResNet18~\cite{brendel2018approximating}           & 4.74$\pm$0.93 & 4.99$\pm$0.53 & 3.57$\pm$0.46 & 7.37$\pm$0.80 & 9.35$\pm$0.69 & 7.77$\pm$0.64 & 10.68$\pm$0.39 & 10.90$\pm$0.72\\
    VGG~\cite{simonyan2014very}              & 1.90$\pm$0.07 & 4.06$\pm$0.23 & 1.05$\pm$0.04 & 2.97$\pm$0.26 & 7.56$\pm$0.69 & 5.73$\pm$0.34 & 4.97$\pm$0.05 & 5.71$\pm$0.18 \\
    BagNet-VGG~\cite{brendel2018approximating}               & 3.52$\pm$0.35 & 4.30$\pm$0.32 & 3.76$\pm$0.33 & 6.32$\pm$0.41 & 8.55$\pm$0.71 & 6.87$\pm$0.70 & 9.26$\pm$0.12 & 9.00$\pm$0.29\\
    Global-Transformer~\cite{dosovitskiy2020image}      & 2.98$\pm$0.46 & 5.78$\pm$0.64 & 1.90$\pm$0.26 & 3.75$\pm$0.34 & 7.77$\pm$0.48& 6.00$\pm$0.54 & 5.06$\pm$0.33 & 7.74$\pm$1.12\\
    Local-Transformer~\cite{vaswani2017attention}      & 2.11$\pm$0.07 & 3.41$\pm$0.18 & 1.41$\pm$0.18 & 4.02$\pm$0.31 & 7.81$\pm$0.57&6.38$\pm$0.42 & 6.14$\pm$0.42 & 8.29$\pm$0.62\\
    Global-Local Transformer& \textbf{1.79$\pm$0.07} & \textbf{3.12$\pm$0.18} & \textbf{0.90$\pm$0.03} & \textbf{2.73$\pm$0.24} & \textbf{6.68$\pm$0.39} &\textbf{5.40$\pm$0.40} & \textbf{4.95$\pm$0.10 }& \textbf{5.68$\pm$0.47}\\
    \bottomrule
    \end{tabular}
\end{table*}

\subsection{Comparison with state-of-the-art neural networks and models of brain age estimation}
Table~\ref{tab:soacompari} shows the comparison with eight state-of-the-art deep networks and the recently published two brain age estimation models in terms of MAE, correlation $r$ and CS ($\alpha$=5 years).
All these models are trained with five-cross validation and the fusion results of the three planes are reported for fair comparison.
We train the SFCN~\cite{peng2019accurate} with 2D and 3D convolutional neural networks, named SFCN 2D and SFCN 3D on the same data.
From Table~\ref{tab:soacompari} we can see that 
(1) the efficient networks (ShuffleNet, SqueezeNet and SFCN 2D) have the largest MAE ($>$ 3.5 years), lowest correlation ($r$ $<$ 0.98) and CS($\alpha$)$<$80\% among algorithms compared.
(2) DenseNet gives better results than other neural networks, including ResNet, WRN and DeepBrainNet. 
 (3) The 3D network of the SFCN provides better results than its 2D version.
(4) Our proposed method outperforms other general-purpose neural networks included in the comparison, and also three networks (SFCN~\cite{peng2019accurate}, DeepBrainNet~\cite{brainwaa160} and FiA-Net~\cite{he2021multi}) specifically designed for brain age estimation.

\begin{table}[!t]
    \centering
    \caption{Comparison with state-of-the-art methods for brain age estimation based on five-cross validation.}
    \label{tab:soacompari}
    \resizebox{0.5\textwidth}{!}{
    \begin{tabular}{l|ccc}
    \toprule
    Method   & MAE & Pearson Correlation ($r$) & CS($\alpha$=5 years)\\
    \midrule
    ShuffleNet V2 (2.0x)~\cite{ma2018shufflenet} & 3.85$\pm$0.12&0.9668$\pm$0.0036&76.90\%$\pm$0.78\\
    SqueezeNet~\cite{iandola2016squeezenet} &  3.71$\pm$0.16&0.9710$\pm$0.0035&77.05\%$\pm$1.53\\
    ResNet50~\cite{he2016deep} & 3.12$\pm$0.08&0.9781$\pm$0.0027&81.88\%$\pm$0.73 \\
    ResNet101~\cite{he2016deep} &   3.15$\pm$0.13&0.9778$\pm$0.0029&81.53\%$\pm$0.98\\
    WRN-50-2~\cite{zagoruyko2016wide} &  3.06$\pm$0.10&0.9786$\pm$0.0028&82.38\%$\pm$0.85\\
    WRN-101-2~\cite{zagoruyko2016wide} &  3.07$\pm$0.10&0.9788$\pm$0.0022&81.97\%$\pm$0.81\\
    DenseNet121~\cite{huang2017densely} &  2.86$\pm$0.08&0.9837$\pm$0.0017&82.87\%$\pm$1.01\\
    DenseNet201~\cite{huang2017densely} &  2.80$\pm$0.07&0.9836$\pm$0.0015&83.72\%$\pm$0.65\\
    \midrule
    *SFCN 2D~\cite{peng2019accurate} (2021)     &   3.58$\pm$0.10 & 0.9754$\pm$0.0023 & 77.73\%$\pm$0.95\\
    *SFCN 3D~\cite{peng2019accurate} (2021)     &   3.04$\pm$0.06&0.9817$\pm$0.0008&81.66\%$\pm$0.50\\
    **FiA-Net$_{fus}$ 3D~\cite{he2021multi} (2021) & 3.00$\pm$0.06 & 0.9840$\pm$0.0000 & 81.75\%$\pm$1.20 \\
    *DeepBrainNet~\cite{brainwaa160} (2020)     & 2.97$\pm$0.11 & 0.9815$\pm$0.0022 & 82.87\%$\pm$0.66\\
    Global-Local Transformer &  \textbf{2.70$\pm$0.03} & \textbf{0.9853$\pm$0.0020} & \textbf{84.53\%$\pm$0.77} \\
    \bottomrule
    \multicolumn{4}{l}{*Models specifically designed for brain age estimation.} \\
    \multicolumn{4}{l}{**Results are directly from the paper.}
    \end{tabular}}
\end{table}

\subsection{Interpretation with multiple patch sizes}
\label{sec:interp}

\begin{figure}[!t]
    \centering
    \includegraphics[width=0.5\textwidth]{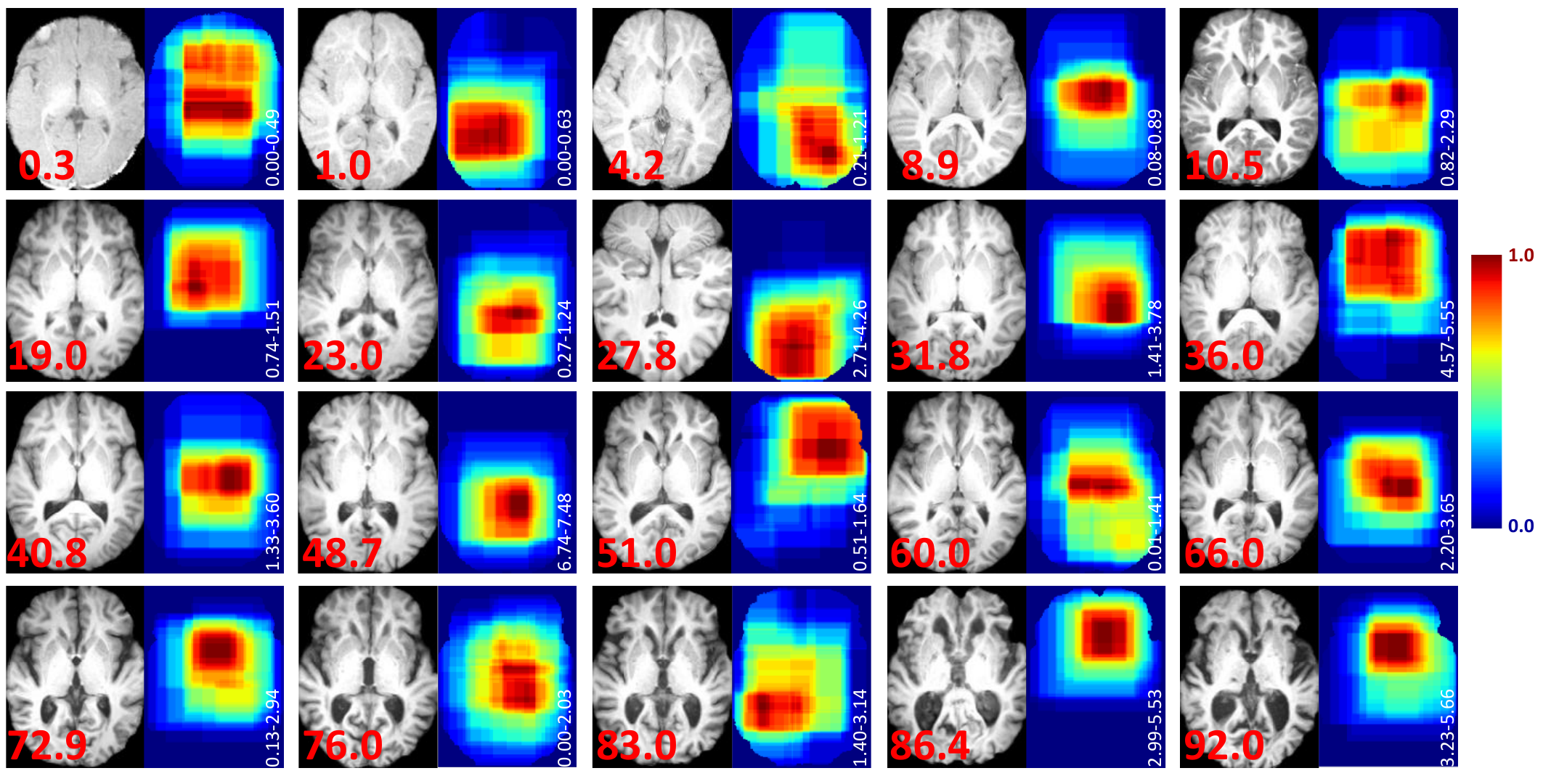}
    \caption{The heatmap (the probability of brain regions which can achieve the lowest MAE) of each brain shows the brain age evidence extracted from local patches. The chronological age is highlighted in red color and the MAE range (minimum-maximum predicted errors) is shown in white color.}
    \label{fig:subjectlevelheatmap}
\end{figure}

In this section, we propose two types of interpretation: subject-level interpretation which highlights the most discriminative patches on each subject and group-level interpretation which shows the most salient brain regions over a group of subjects within a certain age range.
For subject-level interpretation, the 5 patches with the lowest MAEs of each patch size are collected and a heat map is built for visualizing the most informative regions.
For group-level interpretation, we only select the 5 patches on each subject with the lowest MAEs of patch sizes 32 and 40, and then all selected patches from subjects within the age range are averaged to obtain a fine-grained heatmap.
The heatmap shows the probability that the lowest MAE (the best prediction) can be obtained on the brain image.

\begin{figure*}[!t]
    \centering
    \includegraphics[width=0.8\textwidth]{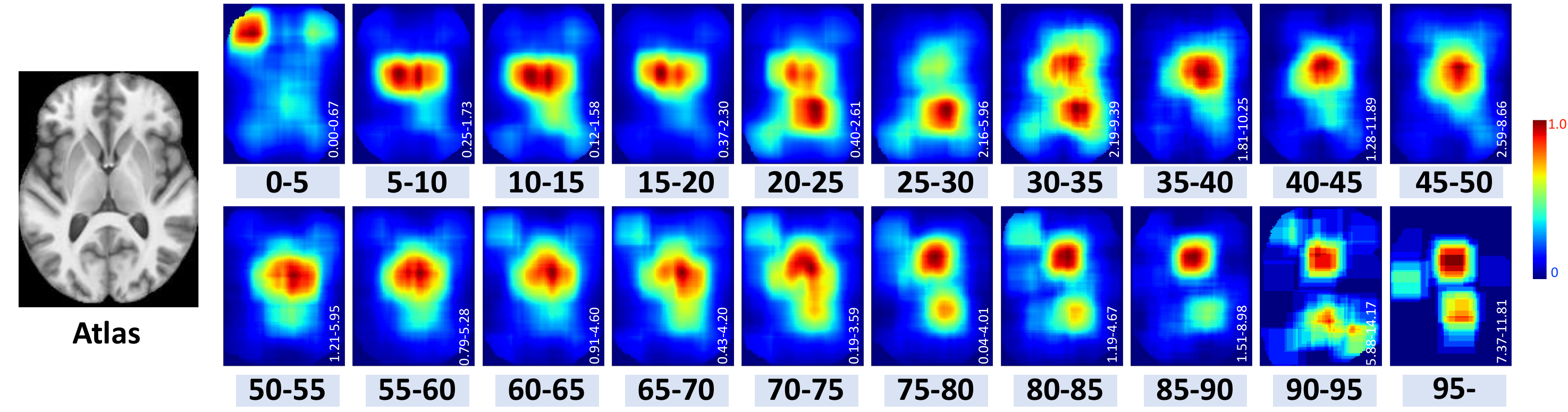}
    \caption{The average heatmaps of subjects on different age ranges. The heatmap shows the probability (red means high probability and blue means low probability, as shown in the color bar on the right) of brain regions which can achieve the lowest MAE on each age group. The MAE range (minimum-maximum predicted errors) is shown in white color.}
    \label{fig:grouplevelheatmap}
\end{figure*}

Fig.~\ref{fig:subjectlevelheatmap} shows the most informative brain regions on each subject computed by averaging the patches with the lowest MAEs with various patch sizes.
 For each brain MRI, the most patches with the lowest MAEs covers the same region, indicating that the salient region (shown in Fig.~\ref{fig:subjectlevelheatmap}) contains the most brain age information than other brain regions. 
In addition, the salient brain age regions are slightly different among subjects with different ages.
To compute the general trends of the salient brain age regions, we average the salient regions on subjects within a certain age range and the results are show in Fig.~\ref{fig:grouplevelheatmap}.
There is a trend of the changes of the salient brain regions over time.
In the children (0-5 years old), the most salient brain age region is on the frontal lobe.
It shifts to the deep gray nuclei region with age range of 5-20 years.
Starting at 20 years, the salient region is gradually shifts to the parietal lobe at 30-35 years old and then shifted back at 35-40 years until 65-70 years.
After 75 years, there are two salient regions which contain the most age information.

\begin{figure}[!t]
    \centering
    \includegraphics[width=0.5\textwidth]{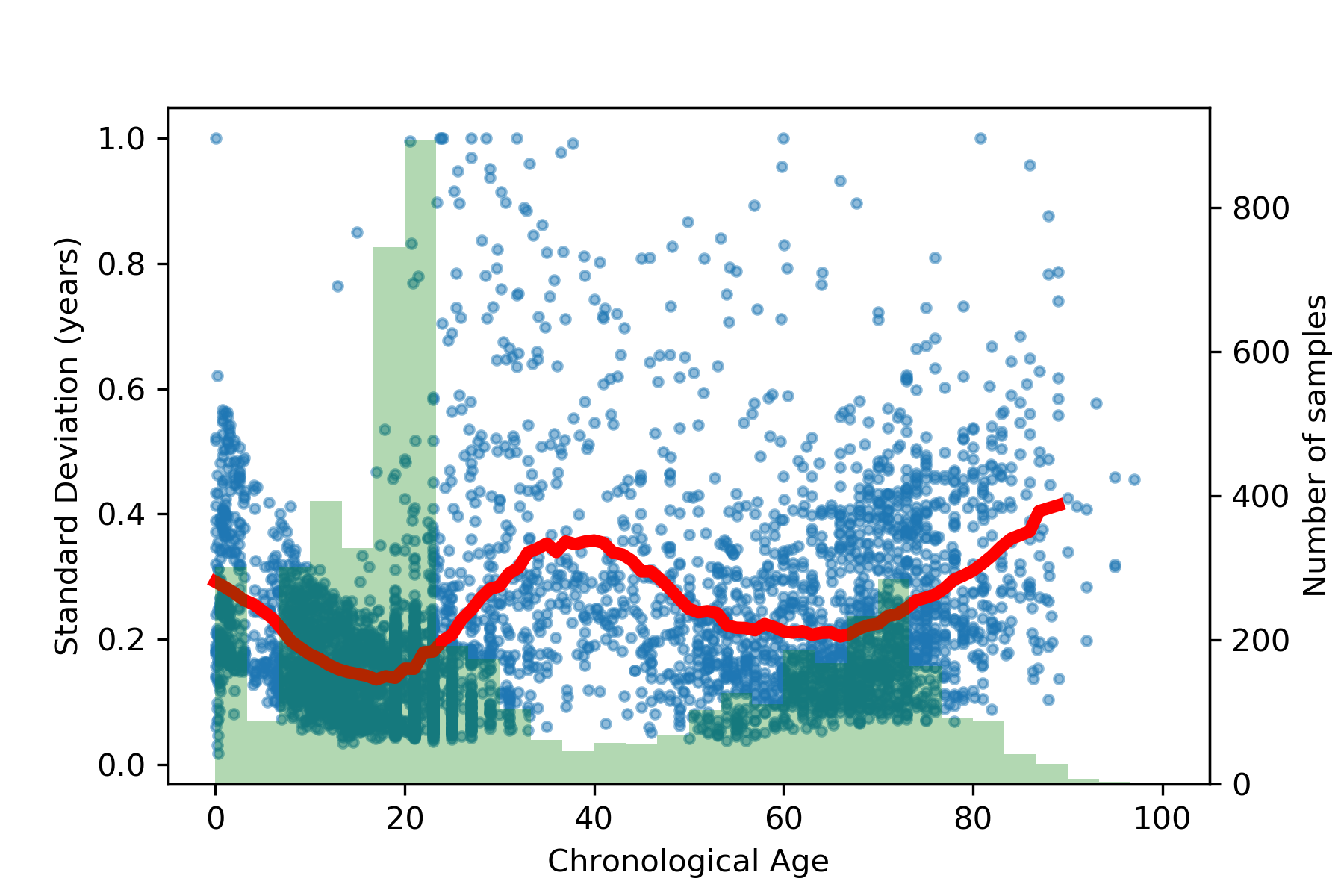}
    \caption{The distribution of the standard deviation $\sigma$ of the brain estimated ages computed by the local patches with various sizes over the years. The red curve is the smooth of the average $\sigma$ on each year, and it comply with the scales in the left $y$ axis. The bar distribution shows the number of samples at each age, following the scales in the right $y$ axis.}
    \label{fig:stdistribution}
\end{figure}

Fig.~\ref{fig:stdistribution} shows the distribution of the standard deviation $\sigma$ (the uncertainty measurement) over lifespan.
A large $\sigma$ means a high differences of the estimating brain age on different brain regions.
It shows that the differences go to lowest on the ages around 20 and 65 years, indicating that the differences among the whole brain regions are smallest on these ages.
Subjects around 40 years of age have the largest differences.
The reason might be that there are fewer training samples on this age range. A similar finding was reported in~\cite{levakov2020deep}.

\begin{figure}
    \centering
    \includegraphics[width=0.5\textwidth]{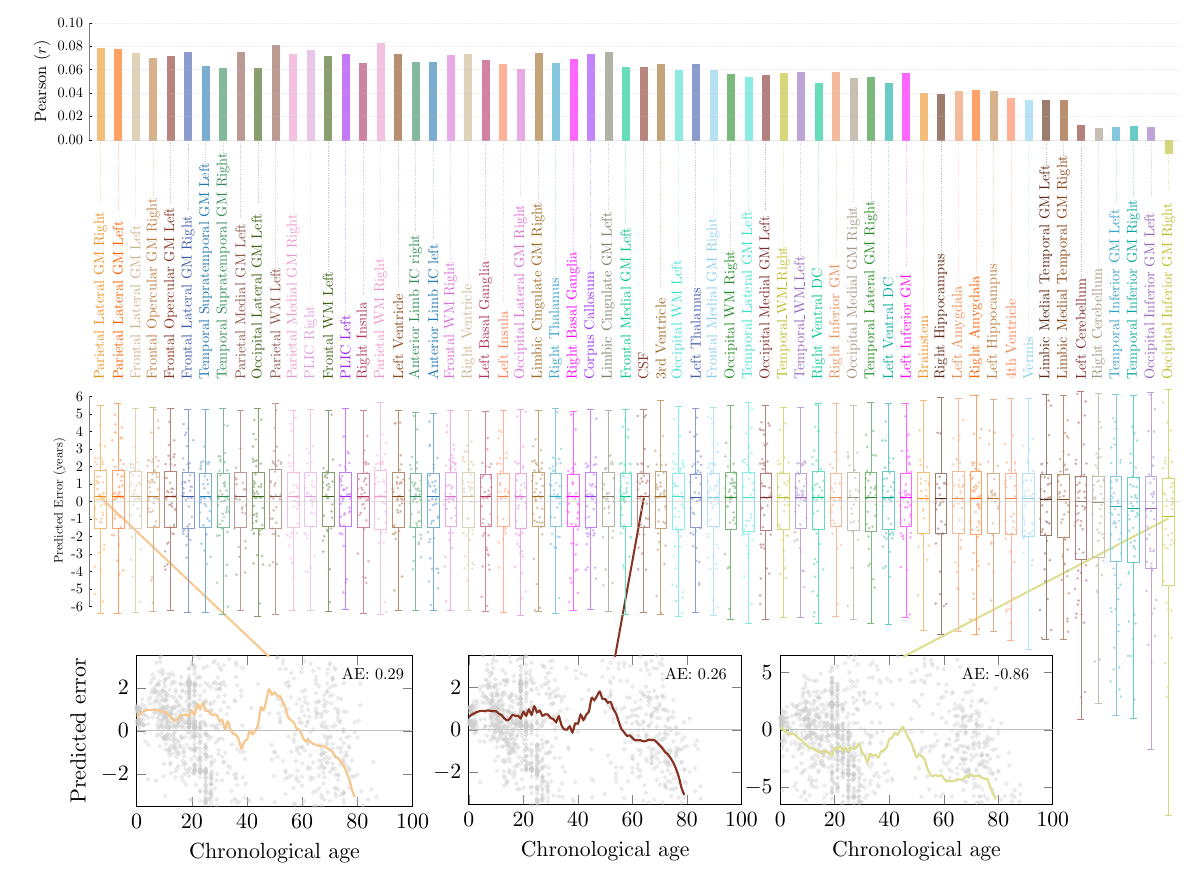}
    \caption{The top bar distribution shows the Pearson correlation between the ICV-normalized region volume and the predicted brain age error on each brain region segmented based on SRI atlas. The box plot of average errors (AE) in years of the predicted brain ages on each auto-segmented brain regions. The regions are sorted in the descending order according the median value.
    The error distributions over lifespan of three different brain regions are shown on the bottom.}
    \label{fig:roiplot}
\end{figure}

The top bar of Fig.~\ref{fig:roiplot} shows the Pearson correlations between the predicted brain age error and the values of intracranial volume (ICV) normalized volume of the brain regions auto-segmented based on the SRI atlas~\cite{rohlfing2010sri24}.
We find that there is no significant correlation existed on the cross-validation data set (n=5,441, $r<$0.1).
The bottom of the Fig.~\ref{fig:roiplot} shows the box plot of each brain region (ROI).
The average errors (AE) of the predicted brain age are slightly different in different brain regions, with the range from 0.29 years of Parietal Lateral GM Right to -0.86 years of Occipital Inferior GM right.
It is also possible to visualize the average errors of each brain regions across the lifespan and three examples are shown in Fig.~\ref{fig:roiplot}.

\section{Discussion and Conclusion}
In this paper, we proposed a novel neural network for brain age estimation called global-local transformer, which optimally fuse the global-context and local detailed information with the attention mechanism. 

We conducted experiments on six public datasets with 5,441 healthy subjects with age of 0-97 years.
By comparing with six different baseline models, the results shown that the proposed global-local transformer provides the best performance for brain age estimation, in terms of MAE, correlation, and cumulative scores with different thresholds (Table~\ref{tab:performace} and Fig.~\ref{fig:CScurve}).
In addition, we also compared the proposed method with eight state-of-the-art neural networks and two specific networks for brain age estimation (Table~\ref{tab:soacompari}).
All of these results have shown that fusing the global-context and local detailed information can improve the performance of brain age estimation on 2D slices.

Our proposed method can also be used for interpreting the evidence for brain age estimation.
We showed the subject-level salient brain regions which provide the lowest MAEs (Fig.~\ref{fig:subjectlevelheatmap}) and the average salient regions over a group of subjects within a certain age range (Fig.~\ref{fig:grouplevelheatmap}). 
It can also be used for computing the differences of brain aging in different brain regions (Fig.~\ref{fig:stdistribution}).
These results demonstrate the advantages of the proposed method which can not only achieve the best performance than other models, but also can visualize the estimated evidence for brain age estimation.

The limitations of the proposed method are summarized as follows:
(1) This study focuses on developing the accurate brain MRI estimation model and MRIs from patients are not involved, similar to other studies in the literature~\cite{feng2020estimating,peng2019accurate,jonsson2019brain}.
Building a machine learning model on healthy cohort with a high performance is the first step to apply it on diseased cohorts, which is our next step.
(2) We also considered the gender information on the last fully-connected layer. We did not report the results since no improvement was obtained. In future, we will investigate where and how to fuse the gender information in the transformer model.
(3) As shown in Table~\ref{tab:soasummary}, the dataset used in our study is not the largest one (the study in~\cite{dinsdale2021learning} used 19,687 subjects with the age of 44-80 years).  
Collecting a large dataset is challenging, especially covering the lifespan 0-100 years of age.
In future, we will continue to collect the dataset to evaluate the performance of the proposed method with different data scales.
(4) As shown in Table~\ref{tab:agegrouperr}, our dataset is unbalanced. The number of samples in 30-60 years is smaller than the number in other age groups, yielding the largest MAEs (5.12 years). 
In future, we will balance the dataset either by re-sampling the training samples~\cite{feng2020estimating} or using data augmentation methods.
(5) It is challenging to fairly compare the proposed method with other studies in the literature since different studies used different datasets, pre-processing, and modalities.
There is no benchmark dataset for brain age estimation.
We summarized the performance of studies in the literature in Table~\ref{tab:soasummary}.
Our method achieved a comparable result (MAE: 2.70 years) which is lower than other studies with lifespan data (covering young and old adults).
(6) Our proposed method can compute the predicted errors on each brain region based on the atlas over the lifespan, as shown in Fig.~\ref{fig:roiplot}. 
However, no significant correlation has been found between the predicted errors and normalized brain volumes on segmented brain regions. 
The results indicate that the mechanism of interpreting the age prediction based on local patches may be different from other indirect interpretation method, such as the explanation maps~\cite{levakov2020deep} computed based on the gradients with the whole brain image as the input.
One future direction is to study the differences between the direct and indirect interpretation methods and their correlations with other natural variabilities in brain morphology.

In conclusion, we have proposed a global-local transformer for brain age estimation using the convolutional neural networks with two pathways: global-pathway to extract global-context information and local-pathway to extract local detailed information. 
The global-context and local detailed information are fused by the attention mechanism. 
The proposed method can achieve the state-of-the-art performance and can highlight the most informative regions.
Future work includes using large and balanced dataset, fusing gender information with MRI and applying the model on patients' MRIs.

\begin{table}[!t]
    \centering
    \caption{Comparison with state-of-the-art methods for brain age estimation based on five-cross validation on the BraTS dataset.}
    \label{tab:soacompariTumor}
    \resizebox{0.5\textwidth}{!}{
    \begin{tabular}{l|ccc}
    \toprule
    Method   & MAE & Pearson Correlation ($r$) & CS($\alpha$=5 years)\\
    \midrule
    ShuffleNet V2 (2.0x)~\cite{ma2018shufflenet} & 8.72$\pm$0.77&0.5642$\pm$0.0979&37.27\%$\pm$5.93\\
    SqueezeNet~\cite{iandola2016squeezenet} &  7.77$\pm$0.58&0.6575$\pm$0.0758&41.20\%$\pm$4.85\\
    ResNet50~\cite{he2016deep} &  7.93$\pm$0.91 &  0.6416$\pm$0.0840 & 42.78\%$\pm$6.56 \\
    ResNet101~\cite{he2016deep} &  7.98$\pm$0.90&0.6272$\pm$0.0693&40.70\%$\pm$8.11\\
    WRN-50-2~\cite{zagoruyko2016wide} &   7.91$\pm$0.76&0.6671$\pm$0.0361&41.47\%$\pm$6.09\\
    WRN-101-2~\cite{zagoruyko2016wide} &  7.94$\pm$0.91&0.6484$\pm$0.0754&42.53\%$\pm$6.92\\
    DenseNet121~\cite{huang2017densely} &  7.11$\pm$0.77&0.7242$\pm$0.0430&46.45\%$\pm$4.70\\
    DenseNet201~\cite{huang2017densely} & 7.34$\pm$0.76&0.6905$\pm$0.0616&44.35\%$\pm$5.58\\
    \midrule
    *SFCN 2D~\cite{peng2019accurate} (2021)    &  10.18$\pm$0.38 & 0.6350$\pm$0.1083 & 32.55\%$\pm$2.63 \\
    *SFCN 3D~\cite{peng2019accurate} (2021) & 9.98$\pm$0.33 & 0.6295$\pm$0.0676 & 32.03\%$\pm$2.64 \\
    *DeepBrainNet~\cite{brainwaa160}(2020)    & 7.90$\pm$0.88 & 0.6816$\pm$0.0463 & 42.52\%$\pm$7.23 \\
    \midrule
    ResNet18~\cite{he2016deep}   &8.02$\pm$0.82&0.6670$\pm$0.0636 & 40.16\%$\pm$6.39\\
    BagNet-ResNet18~\cite{brendel2018approximating}  &8.86$\pm$0.47&0.6400$\pm$0.1004 & 34.38\%$\pm$4.67\\
    VGG~\cite{simonyan2014very}  &11.24$\pm$0.41&0.6066$\pm$0.0708 & 22.83\%$\pm$4.74\\
    BagNet-VGG~\cite{brendel2018approximating} & 7.79$\pm$0.57&0.7023$\pm$0.0743 & 40.93\%$\pm$4.76\\
    Global-Transformer~\cite{dosovitskiy2020image}  & 8.26$\pm$0.79&0.6968$\pm$0.0925 & 36.75\%$\pm$4.97\\
    Local-Transformer~\cite{vaswani2017attention} &  8.06$\pm$0.77&0.6931$\pm$0.0837 & 40.43\%$\pm$5.87\\
    Global-Local Transformer &  \textbf{6.85$\pm$0.65}&\textbf{0.7538$\pm$0.0485} & \textbf{47.78\%$\pm$4.14}\\
    \bottomrule
    \multicolumn{4}{l}{*Models specifically designed for brain age prediction.}
    \end{tabular}}
\end{table}
\begin{figure}[!t]
    \centering
    \includegraphics[width=0.5\textwidth]{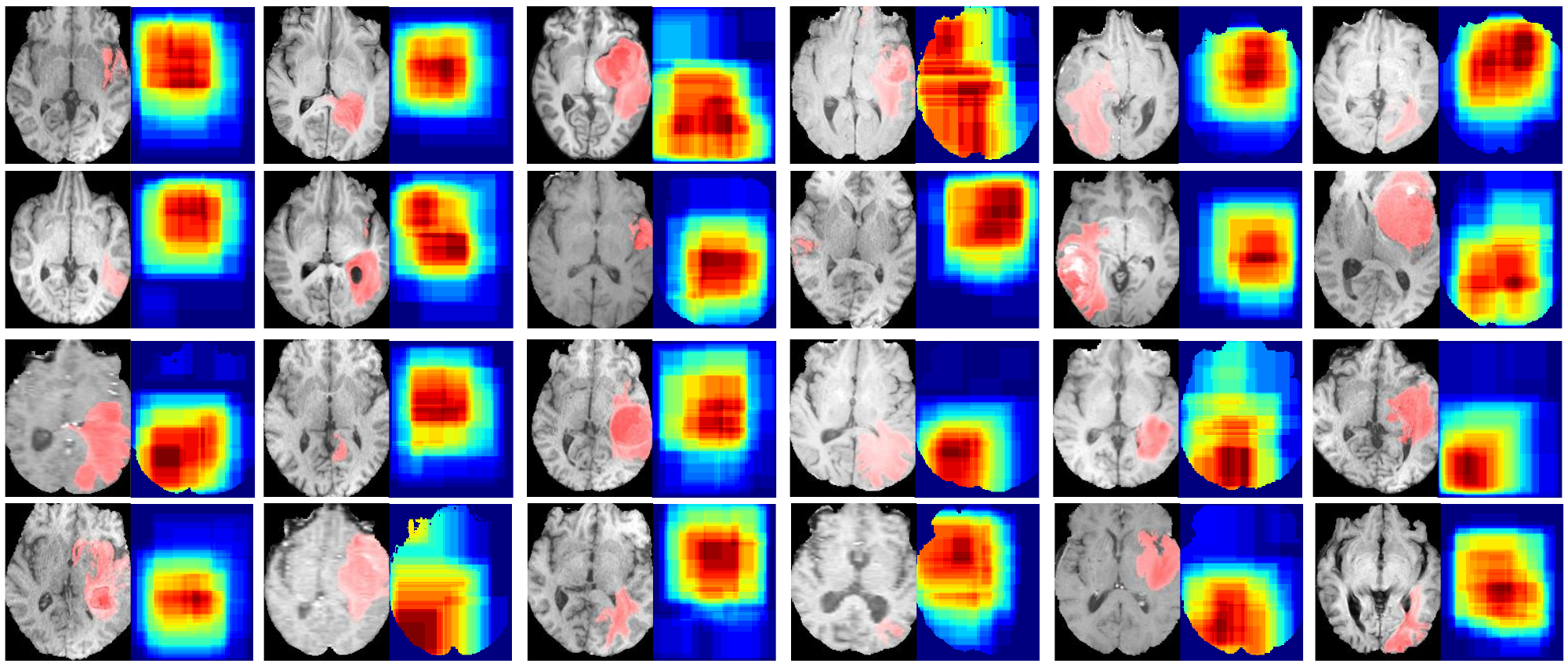}
    \caption{The visualization of the most informative brain regions discovered by the proposed global-local transformer for brain age estimation on subjects with brain tumors. The red mask on the MRIs is the tumor regions labeled by experts.}
    \label{fig:tumorVis}
\end{figure}

\appendix
Our proposed method can be used to estimate brain age on pathology-bearing brain MR images, such as MR images with brain tumor~\cite{menze2014multimodal}.
The biological age of the MRI with tumor is not available and it is only possible to obtain the biological age from subjective evaluation from radiologists which is a time-consuming and subjective processing.
In this paper, we train the machine learning models for the chronological age estimation instead of the biological age estimation.

We collect brain MRIs from BraTS~\cite{menze2014multimodal,bakas2017advancing} and only use subjects whose brain ages are available. 
Finally, there are 382 subjects with age range from 17.4 to 86.6 years.
We concatenate the four modalities: T1w, T1GD, T2w, T2-FLAIR into on images with multiple-channels as input. 
 
 We use the same configurations as the experiments conducted on the healthy cohort: 5 cross-validation is applied on the BraTS dataset and the fusion results of the three planes are reported in this section.
Table~\ref{tab:soacompariTumor} shows the performance of the proposed method, compared with different baseline models, state-of-the-art neural networks and two recently published models for brain age prediction.
Our proposed method achieves the best performance among the seventeen models.
The main reasons are that (1) our proposed method predicts the brain age on local patches. Thus, it can capture the brain age information on the non-tumor brain regions on subjects with tumors;
(2) the global-context information is learnt through the attention, which can automatically find the tumor regions by computing the similarity between healthy and tumor regions, eliminating the affect caused by the tumors.

Similar to Section~\ref{sec:interp}, we also train the global-local transformer with multiple patch sizes and visualize the most informative regions in Fig.~\ref{fig:tumorVis}.
It shows that the salient brain regions does not overlap with the tumor regions, indicating that the predicted age is mainly from the non-tumor brain regions.

Our results show that it may use the brain age estimation for unsupervised brain tumor segmentation in future since the errors of the brain age estimation on tumor and non-tumor regions are different.
The non-tumor regions have the MAE of 7.09$\pm$5.91 years, which is lower than the MAE on tumor regions (8.45$\pm$7.72 years, $p<0.0001$, t-test, two-side).

\section*{Acknowledgment}
This work was funded, in part, by the Harvard Medical School and Boston Children’s Hospital Faculty Development Award (YO), St Baldrick Foundation Scholar Award Grace Fund (YO), and Charles A. King Trust Research Fellowship (SH).

\bibliographystyle{IEEEtran}
\bibliography{IEEEabrv,BrainAge.bib}

\end{document}